\documentclass[10pt,twocolumn]{article}

\usepackage[T1]{fontenc}
\IfFileExists{lmodern.sty}{\usepackage{lmodern}}{}
\IfFileExists{microtype.sty}{\usepackage{microtype}}{}
\usepackage[letterpaper,top=0.75in,bottom=0.85in,left=0.70in,right=0.70in]{geometry}
\usepackage{amsmath}
\usepackage{array}
\usepackage{booktabs}
\usepackage{float}
\usepackage{graphicx}
\usepackage{tabularx}
\usepackage{hyperref}

\hypersetup{
  breaklinks=true,
  colorlinks=true,
  citecolor=blue,
  linkcolor=magenta,
  urlcolor=blue,
  pdfauthor={Tyler Baumgartner, Brandon Tai, Lisa Kaelin-Martin, Candice Fan, Luc Debaupte, Bill Wang, Yi Zhong},
  pdftitle={VoiceCodeBench: Evaluating Exact Structured-Token Recovery in Automatic Speech Recognition}
}
\newcolumntype{Y}{>{\raggedright\arraybackslash}X}

\title{VoiceCodeBench: Evaluating Exact Structured-Token Recovery in Automatic Speech Recognition}
\author{Tyler Baumgartner, Brandon Tai, Lisa Kaelin-Martin, Candice Fan,\\
Luc Debaupte, Bill Wang, and Yi Zhong\\[3pt]
Besimple AI, San Mateo, CA, USA\\
\texttt{\{tyler,yi\}@besimple.ai}}
\date{September 2026}

\begin{document}
\maketitle

\begin{abstract}
Automatic speech recognition is usually evaluated with word error rate (WER),
although voice workflows often require exact written values. VoiceCodeBench
measures whether transcripts preserve identifiers, paths, commands, and other
structured tokens needed by downstream software. It contains 300 human-recorded
English workplace segments (5.59 hours, 85 speakers) and 1,482 audited entities
across 26 types and eight domains. Under a raw-audio-only protocol, we evaluate
19 batch and streaming systems using WER, Canonical Token/Entity Match (CTEM),
and strict segment-level Task Success Rate (TSR). Across systems, WER has little
rank agreement with CTEM (Spearman $\rho=-0.28$) or TSR ($\rho=-0.22$). The best
CTEM and TSR are 91.8\% and 68.7\%. Even the strongest system therefore leaves
nearly one-third of recordings with an unrecovered critical value. Symbol-,
separator-, and boundary-sensitive entities account for most errors.
\end{abstract}


\begin{center}
\begin{minipage}{0.96\columnwidth}
\small
\textbf{Version 2 note.} Version 1 reported ordinary acoustic-reference WER for
12 configurations. This version evaluates 19 configurations with a
format-invariant alternative-reference lattice that accepts the documented
acoustic or canonical rendering at each target span. The resulting system-level
correlations are $-0.28$ for CTEM and $-0.22$ for TSR, replacing the two
$-0.73$ values reported in version 1. The recordings, annotations, CTEM
definition, TSR definition, and recoverability policy are unchanged.
\end{minipage}
\end{center}

\section{Introduction}
\label{sec:introduction}

Speech interfaces increasingly produce software input. A transcript may be
readable yet unusable when it corrupts the identifier, path, amount, or command
that an application must parse, route, store, compare, or execute. Word error
rate (WER) treats word edits largely alike. Losing an article and changing one
digit can contribute similarly, although only the latter may misroute a request
or write an incorrect value.

General ASR corpora measure acoustic and linguistic coverage
\cite{panayotov2015,ardila2020,chen2021}, while application-facing work studies
semantic utility and downstream entities
\cite{wang2003,kim2021,roy2021,szymanski2023}. Related benchmarks evaluate
punctuation and inverse text normalization \cite{meister2023,tan2022}, and
ContextASR-Bench measures recognition with external context \cite{wang2025}.
These strands address transcript fidelity, semantic utility, entity robustness,
normalization, and context use. They leave open the exact recovery of
heterogeneous canonical values from raw audio without hints. This question
matters when punctuation, casing, separators, units, or token boundaries define
a value.

VoiceCodeBench contributes a public, test-only benchmark built around aligned
spoken and canonical entity forms. Its entity-first design provides controlled
coverage of exact-value failure modes. Its raw-audio-only protocol reports CTEM,
strict TSR, and per-type recovery. A versioned verifier exposes evidence and
reasons, and its decisions are independently audited. Across 19 batch and
streaming ASR systems, the study finds weak agreement between WER rankings and
exact-value recovery. It also identifies the entity forms that most often
require safeguards.

\section{Benchmark Design and Collection}
\label{sec:benchmark}

The benchmark contains 300 English segments totaling 5.59 hours from 85
speakers. Its 1,482 target entities span 26 types, eight workplace domains, and
an average of 4.94 targets per recording. Domains cover contact routing,
technical and developer tasks, retail and logistics, finance, health
administration, legal and government work, education and workplaces, and dense
mixed-entity stress cases.

\begin{table}[t]
  \centering
  \small
  \caption{Recording distribution by workflow domain.}
  \label{tab:domains}
  \begin{tabular}{lr}
    \toprule
    Domain & Recordings \\
    \midrule
    Contact/routing & 45 \\
    Technical/IT/developer & 55 \\
    Retail/logistics/order & 45 \\
    Finance/billing & 40 \\
    Healthcare/administration & 35 \\
    Legal/insurance/government & 35 \\
    Education/workplace & 25 \\
    Dense mixed stress & 20 \\
    \bottomrule
  \end{tabular}
\end{table}

\begin{table}[t]
  \centering
  \small
  \setlength{\tabcolsep}{3pt}
  \caption{Entity taxonomy used for structured-token scoring.}
  \label{tab:taxonomy}
  \begin{tabularx}{\columnwidth}{@{}>{\raggedright\arraybackslash}p{0.40\columnwidth}Y@{}}
    \toprule
    Family & Entity types \\
    \midrule
    Contact/routing & Email address, phone number, phone extension, person/team name, postal address \\
    Network/web & URL, IP address, port number \\
    Code/system & Command, CLI flag, file path, environment variable, code symbol, version \\
    Identifiers & Reference ID, product code, account/record number \\
    Numeric/measurement & Currency amount, percentage, measurement, plain number, date, time \\
    Language form & Acronym/initialism, spelled sequence, domain term \\
    \bottomrule
  \end{tabularx}
\end{table}

The six families in Table~\ref{tab:taxonomy} separate values with different
recovery behavior. Conventional numeric forms often tolerate formatting
normalization, while network and code forms can make dots, slashes, dashes,
underscores, and case value-bearing. Construction is entity-first. We allocate a
domain, difficulty, entity count, and entity types, then create unique synthetic
values with spoken and canonical forms. We write a workplace scenario around
that bundle. For example, ``double dash dry dash run'' maps to
\texttt{-{}-dry-run}, and the spoken form must uniquely determine the canonical
value, including any required symbols, case, separators, or units. Generation
was LLM-assisted under fixed metadata and validation constraints. Candidate
items were accepted only after review for domain fit, naturalness, uniqueness,
entity consistency, and recoverability.

Paid contributors recorded the spoken layer and consented to dataset use and
release. Each recording was audited. Files with speech errors or severe quality
problems were rejected and recorded again. Scenarios and sensitive-looking
values are synthetic or reserved, including documentation domains, fictional
NANP 555 numbers, and documentation or private IP ranges. Other identifiers and
addresses are synthetic. Released audio can still identify a voice, so intended
use is ASR evaluation and excludes speaker identification, profiling, and voice
cloning.

VoiceCodeBench is released as a public diagnostic test set. Training on its
labels would undermine test validity, and their public availability precludes a
hidden leaderboard. Reports should disclose model versions, dates, settings,
and any benchmark-specific adaptation. Recordings are assigned light, standard,
dense, or stress difficulty bands. These bands increase transcript length,
entity count, and expected recovery challenge. Their proportions should not be
read as estimates of production frequency.

\begin{table}[t]
  \centering
  \small
  \caption{Difficulty-band composition.}
  \label{tab:difficulty}
  \begin{tabular}{lrrr}
    \toprule
    Band & Recordings & Entities & Words \\
    \midrule
    Light & 30 & 3 & 95--130 \\
    Standard & 114 & 4 & 103--169 \\
    Dense & 93 & 5--6 & 122--177 \\
    Stress & 63 & 7--8 & 156--206 \\
    \bottomrule
  \end{tabular}
\end{table}

Each row includes audio, three transcript layers, target annotations, domain,
difficulty, and coarse speaker and audio-quality metadata. The template,
acoustic, and canonical layers preserve how a value is planned, spoken, and used.
Speakers read the acoustic layer, including explicit symbol, spelling, and case
cues, in a deliberate workplace-dictation style. Acted dialogue and spontaneous
conversation fall outside the collection design. All 300 recordings are
released as one test split to discourage fine-tuning on public labels and
support transparent, repeatable diagnosis.

\section{Evaluation Protocol}
\label{sec:evaluation}

Systems receive only audio bytes. They receive no domain or target metadata,
benchmark-specific prompts, candidate values, custom vocabulary, grammar
constraints, or post-ASR correction. Batch systems process complete files.
Streaming systems receive chronological chunks under a fixed policy, and only
final transcripts are scored. Provider adapters use fixed, documented model,
language, endpoint, and transport settings.

Format-invariant WER lowercases and word-tokenizes each transcript. At every
target span, an acyclic reference lattice accepts either the documented acoustic
rendering or its canonical rendering. Other words use ordinary edit distance.
This prevents \texttt{212-555-0100} from being penalized against ``two one two
five five five zero one zero zero.'' Because punctuation is excluded as a
standalone WER token, WER diagnoses broad transcript quality without measuring
exact written-form recovery.

Entity scoring asks whether the transcript contains enough evidence to recover
each canonical value uniquely. Benign formatting variation is accepted. A
changed digit, unit, symbol, separator, or word that changes the value fails.
Thus, a complete spoken phone-number digit sequence may pass without hyphens. A
missing path suffix or a substitution of milligrams for micrograms fails.
Table~\ref{tab:scoring-examples} illustrates the policy. The verifier assesses
recoverability from available evidence, permitting cosmetic formatting
variation while rejecting value changes and ambiguity.

\begin{table}[t]
  \centering
  \small
  \setlength{\tabcolsep}{2.5pt}
  \caption{Illustrative recoverability decisions.}
  \label{tab:scoring-examples}
  \begin{tabular}{lll}
    \toprule
    Target & Transcript evidence & Decision \\
    \midrule
    \texttt{212-555-0104} & ``212 555 0104'' & Pass: format \\
    \texttt{-{}-dry-run} & ``dry run'' & Fail: symbols \\
    \texttt{.env.provision.local} & ``.env.provision'' & Fail: suffix \\
    \texttt{500 mg} & ``500 micrograms'' & Fail: unit \\
    \bottomrule
  \end{tabular}
\end{table}

\begin{equation}
\mathrm{CTEM}=\frac{N_{\mathrm{correct\ entities}}}{N_{\mathrm{entities}}},
\quad
\mathrm{TSR}=\frac{N_{\mathrm{all\ correct\ recordings}}}{N_{\mathrm{recordings}}}.
\label{eq:metrics}
\end{equation}

CTEM measures value-level recovery. TSR marks a recording successful only when
every target is correct, approximating the fraction of segments that can proceed
without repair. Overall intervals use Wilson's 95\% method. A versioned GPT-5.5
verifier applies the recoverability policy and emits one decision, evidence
span, and reason per target. A 200-decision stratified audit over the original
12-system release found 100\% human agreement, and released prompts and outputs
enable policy inspection and re-scoring. The baseline suite contains 19
commercial-API and open-model configurations evaluated in batch or streaming
mode.

Spearman correlations are computed over the 19 system-level aggregate scores.
These system-level correlations summarize agreement between rankings. They do
not measure recording-level associations or establish that changing WER causes
a change in CTEM. Each prediction artifact records its model identifier and
transcript, while repository documentation records provider settings. Because
commercial endpoints may change over time, their results remain snapshots.

\section{Results}
\label{sec:results}

Table~\ref{tab:results} reports the current 19-system comparison. WER ranges from
4.3\% to 57.3\%, CTEM from 33.9\% to 91.8\%, and TSR from 6.3\% to 68.7\%.

\begin{table*}[t]
  \centering
  \small
  \setlength{\tabcolsep}{2.8pt}
  \caption{Raw-audio-only results on 300 recordings and 1,482 entities. All
  values are percentages. Best values are bold.}
  \label{tab:results}
  \begin{tabular}{lrrr@{\hspace{8mm}}lrrr}
    \toprule
    \multicolumn{4}{c}{Batch} & \multicolumn{4}{c}{Streaming} \\
    \cmidrule(r){1-4}\cmidrule(l){5-8}
    System & WER $\downarrow$ & CTEM $\uparrow$ & TSR $\uparrow$ &
    System & WER $\downarrow$ & CTEM $\uparrow$ & TSR $\uparrow$ \\
    \midrule
    ElevenLabs Scribe v2 & 6.9 & 91.6 & 67.7 & OpenAI gpt-live-transcribe & 5.9 & \textbf{91.8} & \textbf{68.7} \\
    Deepgram Nova-3 & 8.2 & 90.9 & \textbf{68.7} & Cartesia Ink 2 & 9.3 & 90.1 & 63.7 \\
    OpenAI gpt-transcribe & 6.1 & 89.3 & 61.0 & Deepgram Nova-3 & 9.3 & 88.9 & 61.7 \\
    Google Cloud Chirp 3 & 5.8 & 88.8 & 60.3 & Meta Muse Voice Transcribe 1.0 & 7.4 & 86.8 & 55.7 \\
    Whisper large-v3 & 5.8 & 87.6 & 54.3 & Google Cloud Chirp 3 & 6.7 & 86.1 & 50.3 \\
    Meta Muse Voice Transcribe 1.0 & 7.5 & 87.6 & 55.3 & ElevenLabs Scribe v2 Realtime & 6.8 & 84.0 & 46.3 \\
    AssemblyAI Universal-3 Pro & \textbf{4.3} & 84.5 & 50.3 & AssemblyAI Universal-3 Pro & 6.1 & 78.3 & 33.0 \\
    Inkling-NVFP4 via Modal & 4.9 & 84.3 & 49.7 & Amazon Transcribe & 10.3 & 75.2 & 33.7 \\
    Inworld STT 1 & 9.7 & 83.7 & 44.0 & Inworld STT 1 & 57.3 & 33.9 & 6.3 \\
    NVIDIA Parakeet TDT 0.6B v3 & 9.3 & 78.6 & 37.3 & & & & \\
    \bottomrule
  \end{tabular}
\end{table*}

Transcript and entity rankings diverge. Across the 19 systems, Spearman
correlation is $\rho=-0.28$ between WER and CTEM and $\rho=-0.22$ between WER
and TSR, indicating only weak monotonic association in this sample.

\begin{figure*}[t]
  \centering
  \includegraphics[width=0.72\textwidth]{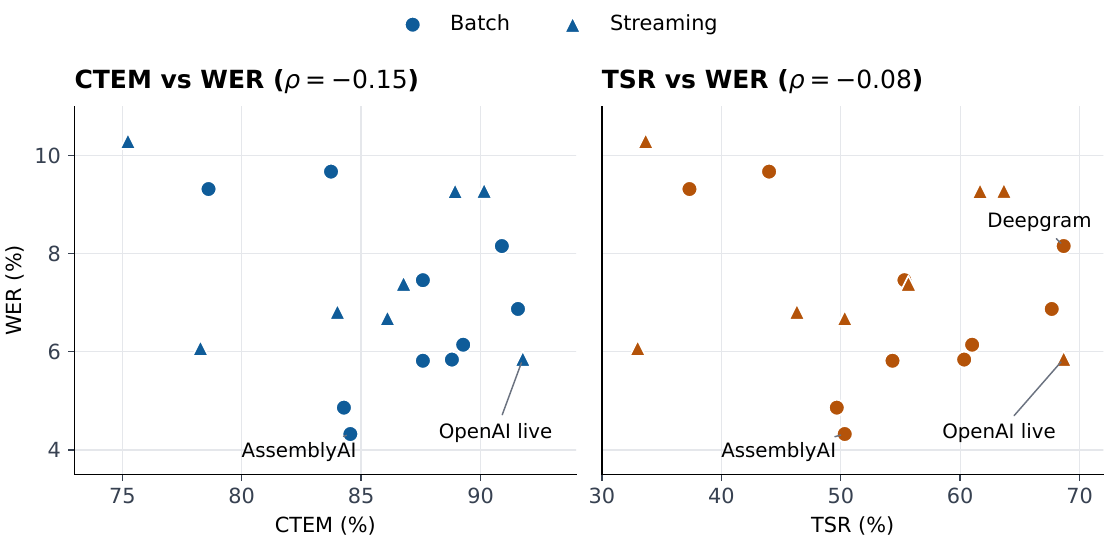}
  \caption{WER against CTEM and TSR for 18 configurations. Inworld STT 1
  streaming is omitted for scale. Plot correlations use the displayed points.
  Lower WER and higher CTEM/TSR are better.}
  \label{fig:wer-entity-scatter}
\end{figure*}

Figure~\ref{fig:wer-entity-scatter} omits Inworld STT 1 streaming for scale. The
excluded system remains in Table~\ref{tab:results} and all 19-system analyses.
Among the displayed configurations, Spearman $\rho$ is $-0.15$ for CTEM and
$-0.08$ for TSR, so the rank mismatch remains without the outlier. AssemblyAI
Universal-3 Pro batch has the lowest WER but ranks twelfth on CTEM, whereas
OpenAI gpt-live-transcribe has the best CTEM and ties Deepgram Nova-3 batch for
the best TSR, although neither has the lowest WER.

The leading systems have overlapping intervals. OpenAI gpt-live-transcribe has CTEM 95\% CI
90.3--93.1 and TSR CI 63.2--73.7. Deepgram Nova-3 has the same TSR interval,
while ElevenLabs Scribe v2 reaches CTEM 91.6\% and TSR 67.7\%. Even the best TSR
leaves 31.3\% of segments with at least one unrecovered target, an operationally
important gap because one corrupted entity can invalidate an otherwise fluent
transcript.

Across 28,158 system--entity decisions, 4,713 fail. URLs, commands, email
addresses, file paths, and postal addresses contribute 2,426 failures, or 51.5\%
of the total. Median CTEM is lowest for URLs (46.8\%), commands (52.0\%), file
paths (54.9\%), postal addresses (55.0\%), and email addresses (63.1\%). These
forms depend on symbols, separators, and exact boundaries, and
Table~\ref{tab:hard-types} shows that their medians conceal large system spreads.
They therefore discriminate among systems whose general transcript scores may
be close.

\begin{table}[t]
  \centering
  \small
  \caption{CTEM distribution for the five hardest entity types.}
  \label{tab:hard-types}
  \begin{tabular}{lrrrr}
    \toprule
    Type & Targets & Min. & Median & Max. \\
    \midrule
    URL & 62 & 9.7 & 46.8 & 66.1 \\
    Command & 50 & 8.0 & 52.0 & 78.0 \\
    File path & 51 & 7.8 & 54.9 & 70.6 \\
    Postal address & 40 & 27.5 & 55.0 & 77.5 \\
    Email address & 65 & 9.2 & 63.1 & 80.0 \\
    \bottomrule
  \end{tabular}
\end{table}

Aggregated by family, numeric/measurement reaches 93.7\% CTEM and language form
reaches 92.1\%. Network/web and code/system trail at 65.9\% and 69.1\%,
respectively.

\begin{table}[t]
  \centering
  \small
  \caption{Aggregate CTEM by entity family across 19 systems.}
  \label{tab:families}
  \begin{tabular}{lrr}
    \toprule
    Family & Targets & CTEM (\%) \\
    \midrule
    Numeric/measurement & 400 & 93.7 \\
    Language form & 202 & 92.1 \\
    Identifiers & 305 & 87.3 \\
    Contact/routing & 213 & 75.3 \\
    Code/system & 245 & 69.1 \\
    Network/web & 117 & 65.9 \\
    \bottomrule
  \end{tabular}
\end{table}

\section{Discussion}
\label{sec:discussion}

\subsection{Operational Interpretation}

WER remains useful for broad transcript quality. CTEM and TSR extend evaluation
to software-facing speech. CTEM estimates individual value-repair load,
TSR estimates the share of complete segments that could proceed without any
entity repair under this benchmark's entity distribution, and per-type recovery
identifies where confirmation, typed validation, constrained decoding, or human
review may be needed.

TSR is intentionally sensitive to entity load. Pooled across all 19 systems,
CTEM is 83.6\%, 81.6\%, 83.3\%, and 84.8\% from light through stress, whereas
TSR falls from 63.5\% to 52.7\%, 51.2\%, and 41.4\%. The nearly stable entity
accuracy but declining segment success reflects the greater chance that at least
one entity fails in a denser segment. Neither metric weights errors by
consequence. A wrong dosage, malformed test path, and mistyped routing name each
count as one entity error although their real-world costs differ. Deployment
evaluations should therefore interpret TSR with CTEM, per-type scores, the
controlled difficulty mix, and domain-specific risk.

The raw-audio-only scores provide a deployment baseline. Applications may
improve on them with candidate lists, domain context, confirmation turns, typed
parsers, range checks, or human review, then report those conditions separately.
The failure concentration points to targeted controls. URLs, paths, commands,
and email addresses benefit from character-level confirmation, while amounts
and units benefit from typed validation before a workflow commits an action.

Batch and streaming entries describe the evaluated endpoints. The comparison
does not isolate latency or streaming effects. Models, decoding behavior, and
endpoint implementations can differ, so cross-mode gaps should not be attributed
to streaming alone. The benchmark scores transcript evidence as a proxy for
workflow readiness. Executed downstream actions fall outside the evaluation,
and application parsers may accept different surface forms.

The system-selection consequence is concrete. Optimizing only WER selects the
4.3\%-WER batch system. The CTEM/TSR leader improves those scores by 7.2 and 18.3
points, respectively, while increasing WER by only 1.5 points. That choice should
follow application cost. A note-taking tool may prioritize readability, while a
system that writes identifiers or executes commands may prefer structured-token
recovery even at a modest WER penalty.

VoiceCodeBench also supports regression diagnosis. Because each target retains
its type, acoustic form, and canonical form, a release can reveal a local decline
in path or URL handling even when aggregate WER appears stable.

\subsection{Limitations}

VoiceCodeBench is English-only and uses deliberate workplace dictation with
synthetic scenarios. It does not represent all spontaneous, overlapping, noisy,
long-form, multilingual, or domain-specific speech. The 85-speaker sample is
insufficient for fine-grained subgroup claims. Coarse speaker labels are
descriptive, and the study lacks the design and statistical power to rank
systems by demographic group.

Recoverability scoring depends on a documented policy and an LLM verifier. The
human audit supports this implementation, although borderline cases still
require judgment. The 200-decision audit sampled the original 12-system suite.
It therefore covers only part of the expanded 19-system release. Perfect
agreement on this sample cannot establish a zero verifier error rate.

Several of the 19 configurations share providers or model families, creating
dependence within the ASR sample. The rank correlations are descriptive, and a
future release with more systems may change their magnitude. Wilson intervals
describe uncertainty for the reported proportions but do not capture clustering
by speaker, recording, or entity type. Paired or cluster-aware resampling would
better support formal comparisons between close systems.

Public labels permit adaptation, and provider systems can drift. Reported
scores are therefore dated, reproducible snapshots of evaluated model
configurations. Reports should pair dated, versioned prediction artifacts with
WER, CTEM, TSR, and per-type results, keeping raw and adapted runs separate so
the effect of prompts, context, or repair logic remains measurable. Synthetic
values improve control and reduce exposure of real operational data, but their
frequency and context may differ from production. Comparative diagnosis should
therefore be followed by testing on the intended application. Future work should
add languages, spontaneous speech, channel noise, and application context, while
deterministic canonicalizers could reduce verifier dependence under documented
edge-case rules.

\subsection{Conclusion}

VoiceCodeBench measures whether readable ASR output preserves the exact values
required by a workflow, a narrow but consequential failure mode. Evaluating this
failure mode requires metrics beyond WER. Together, WER, CTEM, TSR, and per-type
recovery distinguish transcript quality, value-level repair burden,
complete-segment success, and concentrated risk, turning ASR evaluation into a
more actionable deployment decision. The dataset, annotations, scoring code,
prompts, baseline outputs, and detailed documentation are available in the
\href{https://huggingface.co/datasets/besimple-ai/voice-code-bench}{project repository}.

\section*{Acknowledgments}

This work was funded by Besimple, Inc. (Besimple AI).

\begingroup
\sloppy

\endgroup

\clearpage
\onecolumn
\appendix

\section{Detailed Benchmark Composition}
\label{app:composition}

Table~\ref{tab:entity-counts} gives the full entity-type distribution. The
counts sum to 1,482 and are fixed by the released annotations.

\begin{table}[H]
  \centering
  \scriptsize
  \caption{Target-entity counts by entity type.}
  \label{tab:entity-counts}
  \begin{tabularx}{\textwidth}{@{}YrYrYr@{}}
    \toprule
    Entity type & Count & Entity type & Count & Entity type & Count \\
    \midrule
    \texttt{reference\_id} & 150 & \texttt{url} & 62 & \texttt{code\_symbol} & 35 \\
    \texttt{spelled\_sequence} & 97 & \texttt{measurement} & 61 & \texttt{environment\_variable} & 35 \\
    \texttt{product\_code} & 90 & \texttt{phone\_number} & 60 & \texttt{phone\_extension} & 30 \\
    \texttt{date} & 89 & \texttt{time} & 60 & \texttt{port\_number} & 30 \\
    \texttt{acronym\_or\_initialism} & 85 & \texttt{file\_path} & 51 & \texttt{version} & 30 \\
    \texttt{currency\_amount} & 75 & \texttt{command} & 50 & \texttt{ip\_address} & 25 \\
    \texttt{account\_or\_record\_number} & 65 & \texttt{percentage} & 50 & \texttt{domain\_term} & 20 \\
    \texttt{plain\_number} & 65 & \texttt{cli\_flag} & 44 & \texttt{person\_or\_team\_name} & 18 \\
    \texttt{email\_address} & 65 & \texttt{postal\_address} & 40 & & \\
    \bottomrule
  \end{tabularx}
\end{table}

Table~\ref{tab:diagnostic-slices} reports descriptive domain and difficulty
slices pooled across the 19 evaluated configurations. CTEM pools entity
decisions within each slice, and TSR pools configuration-recording outcomes.
The slices support diagnosis but do not estimate production frequencies or
independent provider effects.

\begin{table}[H]
  \centering
  \scriptsize
  \begin{minipage}[t]{0.57\textwidth}
    \centering
    \textbf{(a) Workflow domain}\\[3pt]
    \begin{tabular}{lrrrr}
      \toprule
      Domain & Rec. & Ent. & CTEM & TSR \\
      \midrule
      Contact/routing & 45 & 213 & 80.3 & 40.9 \\
      Technical/IT/developer & 55 & 260 & 69.8 & 27.3 \\
      Retail/logistics/order & 45 & 214 & 81.9 & 41.6 \\
      Finance/billing & 40 & 190 & 82.6 & 48.8 \\
      Healthcare/administration & 35 & 167 & 93.8 & 81.1 \\
      Legal/insurance/government & 35 & 167 & 91.6 & 74.4 \\
      Education/workplace & 25 & 119 & 89.5 & 68.0 \\
      Dense mixed stress & 20 & 152 & 87.6 & 48.7 \\
      \bottomrule
    \end{tabular}
  \end{minipage}
  \hfill
  \begin{minipage}[t]{0.39\textwidth}
    \centering
    \textbf{(b) Difficulty band}\\[3pt]
    \begin{tabular}{lrrrr}
      \toprule
      Band & Rec. & Ent. & CTEM & TSR \\
      \midrule
      Light & 30 & 90 & 83.6 & 63.5 \\
      Standard & 114 & 456 & 81.6 & 52.7 \\
      Dense & 93 & 479 & 83.3 & 51.2 \\
      Stress & 63 & 457 & 84.8 & 41.4 \\
      \bottomrule
    \end{tabular}
  \end{minipage}
  \caption{Pooled CTEM and TSR percentages by workflow domain and difficulty
  band across all 19 evaluated configurations. Rec. and Ent. give the numbers
  of recordings and annotated entities in each slice.}
  \label{tab:diagnostic-slices}
\end{table}

\section{Generation Workflow}
\label{app:generation}

Transcript and entity generation used a repository-aware LLM workflow under
fixed metadata, taxonomy, and validation constraints. The workflow preserved
each segment's domain, scenario, difficulty, entity allocation, acoustic and
canonical annotations, and three transcript layers. Deterministic checks tested
schema validity, uniqueness, coverage, and alignment before acceptance. Human
review then assessed domain fit, naturalness, entity consistency, and whether
the spoken evidence uniquely determined every canonical value.

\section{Verifier Prompt}
\label{app:verifier-prompt}

The versioned verifier artifact \texttt{openai\_gpt\_5\_5\_v1} uses the system
prompt reproduced below. The user message supplies the datapoint identifier,
target entities, and ASR transcript as JSON. The response is constrained by the
released strict JSON schema. The repository copy remains the authoritative
machine-readable artifact.

\begingroup
\small
\begin{verbatim}
You verify whether each gold Voice Code Bench entity is present in a
raw speech-to-text transcript.

Return only valid JSON matching the provided schema.

Entity types:
- email_address: full email addresses, including dots, hyphens,
  underscores, plus tags, and spoken at/dot separators.
- phone_number: dialable phone numbers. The canonical form uses
  XXX-XXX-XXXX for US numbers in this dataset.
- phone_extension: phone extension values such as ext74 or ext4821.
- person_or_team_name: unambiguous public, team, organization, or
  routing names whose written form is recoverable from the transcript.
- postal_address: mailing addresses, suites, floors, units, cities,
  state abbreviations, and ZIP-like values.
- url: web URLs and hostnames, including subdomains, paths, and query
  strings.
- ip_address: IPv4 addresses in dotted decimal form.
- port_number: network port numbers.
- command: literal CLI commands or command snippets whose exact tokens
  matter.
- cli_flag: command-line flags such as --dry-run, --config, or -k.
- file_path: file paths, directory paths, filenames, hidden files, and
  extensions.
- environment_variable: environment variable names such as DATABASE_URL
  or NODE_ENV.
- code_symbol: function names, class names, package names, branch names,
  config keys, and identifiers.
- version: software, firmware, API, schema, or model versions.
- reference_id: cases, tickets, claims, invoices, appointments,
  confirmations, tracking numbers, and other operational IDs.
- product_code: SKUs, serials, model numbers, part numbers, lot numbers,
  and device identifiers.
- account_or_record_number: account numbers, masked account tails,
  medical record numbers, member IDs, and record locators.
- currency_amount: monetary amounts with an explicit currency.
- percentage: percentages, rates, APRs, tax rates, discounts, and
  allocation percentages.
- measurement: numeric values with units, including dosage, weight,
  length, volume, temperature, pressure, duration, and lab values.
- plain_number: exact standalone numbers without a unit or currency.
- date: calendar dates where the exact date matters.
- time: appointment times, deadlines, time windows, and time zones.
- acronym_or_initialism: spoken or letter-by-letter acronyms and
  initialisms, including conventional punctuation.
- spelled_sequence: values explicitly spelled letter-by-letter,
  including names and mixed letter/digit sequences.
- domain_term: specialized vocabulary that matters for task success and
  is not covered by a more structured type.

Rules:
- Use only the STT transcript.
- Check every target entity independently and return one result for each
  target_index.
- Use the target acoustic field as the expected spoken form and the
  canonical field as the exact value to recover.
- Mark present true only when the transcript contains enough evidence to
  recover that exact canonical value.
- Accept casing, punctuation, spacing, and formatting differences only
  when the same value remains recoverable.
- Reject wrong, missing, extra, or substituted letters, digits,
  separators, units, dates, times, amounts, or words that change the
  target value.
- Copy target_index, type, and canonical exactly from the target entity
  input.
- Set present true only when the target is supported by the transcript.
- Include an evidence field with the exact transcript substring that
  supports the decision. For absent entities, use the closest corrupted
  substring when available, otherwise use an empty string.
- Include a short reason explaining the present/absent decision.
- Do not return entities that are not listed as target entities.
- Do not explain outside the JSON.
\end{verbatim}
\endgroup

\end{document}